\documentclass[sigconf]{acmart}

\usepackage{stfloats}
\usepackage{makecell}
\usepackage{soul}
\AtBeginDocument{%
  \providecommand\BibTeX{{%
    \normalfont B\kern-0.5em{\scshape i\kern-0.25em b}\kern-0.8em\TeX}}}

\acmSubmissionID{1515}

\copyrightyear{2024}
\acmYear{2024}
\setcopyright{acmlicensed}\acmConference[HRI '24]{Proceedings of the 2024
ACM/IEEE International Conference on Human-Robot Interaction}{March
11--14, 2024}{Boulder, CO, USA}
\acmBooktitle{Proceedings of the 2024 ACM/IEEE International Conference on
Human-Robot Interaction (HRI '24), March 11--14, 2024, Boulder, CO, USA}
\acmDOI{10.1145/3610977.3634937}
\acmISBN{979-8-4007-0322-5/24/03}

\begin{document}

\title{Making Informed Decisions: Supporting Cobot Integration Considering Business and Worker Preferences}

\author{Dakota Sullivan}
\affiliation{
 \institution{University of Wisconsin-Madison}
 \city{Madison}
 \state{Wisconsin}
 \country{USA}
 \postcode{53706}
}
\email{dsullivan8@wisc.edu}

\author{Nathan Thomas White}
\affiliation{
 \institution{University of Wisconsin-Madison}
 \city{Madison}
 \state{Wisconsin}
 \country{USA}
 \postcode{53706}
}
\email{ntwhite@wisc.edu}

\author{Andrew Schoen}
\affiliation{
 \institution{University of Wisconsin-Madison}
 \city{Madison}
 \state{Wisconsin}
 \country{USA}
 \postcode{53706}
}
\email{schoen@cs.wisc.edu}

\author{Bilge Mutlu}
\affiliation{
 \institution{University of Wisconsin-Madison}
 \city{Madison}
 \state{Wisconsin}
 \country{USA}
 \postcode{53706}
}
\email{bilge@cs.wisc.edu}

\begin{abstract}
Robots are ubiquitous in small-to-large-scale manufacturers. While collaborative robots (cobots) have significant potential in these settings due to their flexibility and ease of use, proper integration is critical to realize their full potential. Specifically, cobots need to be integrated in ways that utilize their strengths, improve manufacturing performance, and facilitate use in concert with human workers. Effective integration requires careful consideration and the knowledge of roboticists, manufacturing engineers, and business administrators. We propose an approach involving the stages of \textit{planning}, \textit{analysis}, \textit{development}, and \textit{presentation}, to inform manufacturers about cobot integration within their facilities prior to the integration process. We contextualize our approach in a case study with an SME collaborator and discuss insights learned.

\end{abstract}



\settopmatter{printacmref=false}
\setcopyright{none}
\renewcommand\footnotetextcopyrightpermission[1]{}
\pagestyle{plain}

\maketitle

\begin{figure}
  \centering
  \includegraphics[width=\linewidth, height=14.2cm]{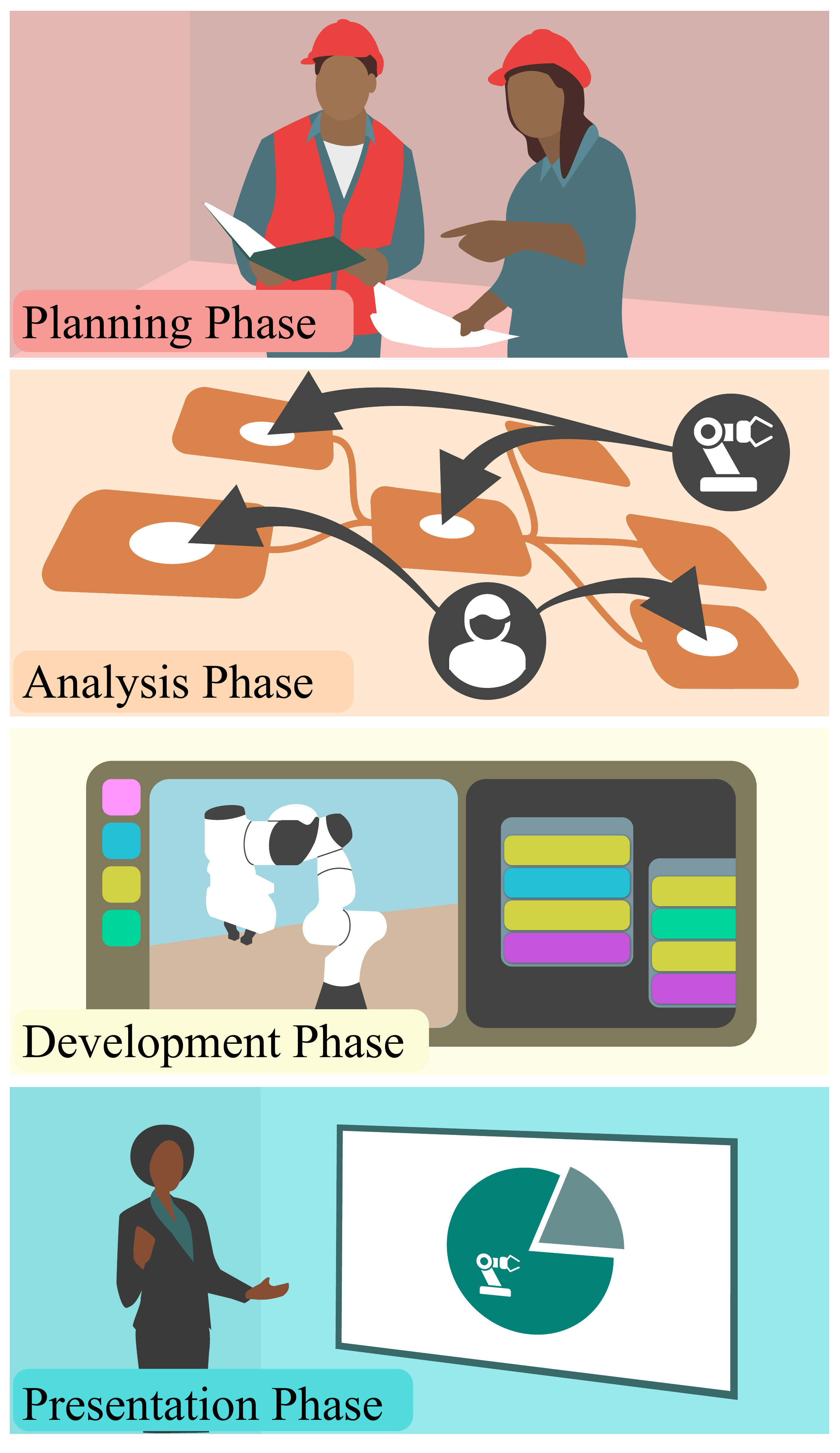}
  \vspace*{-2mm}
  \caption{A depiction of the four-phase cobot integration approach proposed within this paper: planning for integration, analyzing workflows, developing simulations, and presenting to the manufacturer.}
  \vspace*{-2mm}
  \label{fig:teaser}
\end{figure}

\section{Introduction} 

Cobots are increasingly utilized across various tasks and domains \cite{javaid2022significant} and hold particular potential within manufacturing settings \cite{liu2022application}. This potential comes from the versatility and flexibility that cobots provide, as they are relatively easy to reprogram and repurpose without an integrator. Given their ability to work in conjunction with human workers and perform precise, repetitive actions, cobots possess a skill set that makes them very effective in tasks such as assembly, palletizing, packaging, kitting, and tool use for caulking, gluing, and sanding. However, most prior efforts to analyze cobot integration have focused on the associated engineering challenges that emerge following the choice to integrate.

Although there has been significant progress in the development of technical approaches to integration, several key questions remain: ``\textit{How can a cobot complement existing human-only work processes};'' ``\textit{are there subtasks that human workers prefer the cobot perform};'' and ``\textit{is it possible to integrate a cobot while maintaining these preferences?}'' These questions are important to address to ensure that a cobot is operating safely, being utilized effectively, meeting worker preferences, and has a positive impact on business outcomes, as these factors affect cobot adoption \cite{simoes2019drivers, silva2022criteria, berx2022examining, aaltonen2019experiences}. Additionally, as these questions are not fully considered in the existing integration paradigm, organizational leaders, including owners and managers of small and mid-size enterprises (SMEs), may lack the knowledge or understanding required to make informed decisions about cobot integration in their workspaces. When cobots are integrated under these circumstances, the result may be poor utilization of their collaborative capabilities, disruption in existing worker processes, and only partial realization of potential improvements in overall business outcomes \cite{michaelis2020collaborative, paliga2022human}.

In this work, we focus on the integration process, examine decisions that occur prior to implementation, and propose an approach to collaborating with manufacturers. Our proposed approach includes four phases: \textbf{planning} for integration, \textbf{analysis} of existing workflows, \textbf{development} of new human-cobot workflows, and \textbf{presentation} of results to stakeholders (Figure \ref{fig:teaser}). This process allows stakeholders within manufacturing settings to make informed decisions about cobot integration, address questions related to worker and business preferences, and consider practical engineering constraints. To illustrate our approach, we discuss each phase within the context of our collaboration with an SME manufacturer. Following this discussion, we examine feedback from our collaborator.

\section{Background} 

\subsection{Cobot Usage}

SMEs are increasingly using cobots in their processes, in part due to their marketed usability and benefits for collaboration \cite{simoes2019drivers, simoes2020factors}, and the potential for reduction in cycle time of their processes \cite{enrique2021advantages}. Cobots can help to reduce repetitive tasks for operators \cite{marvel2014collaborative} and assist them in their tasks, such as by holding objects the operator is working on \cite{munzer2018efficient}. This practice of the cobot assisting operators is well explored within the research community \cite{peshkin1999cobots, colgate2003intelligent, bi2021safety}.

However, prior work has noted that the usage of cobots by SMEs has primarily been as a cheaper alternative to traditional manufacturing robots, resulting in SMEs not fully utilizing their collaborative capabilities \cite{michaelis2020collaborative, guertler2023robot, wallace2021getting}. In part, this under-utilization of the collaborative aspect of cobots can be attributed to the difficulty of finding appropriate tasks applications, misunderstanding how to utilize cobots effectively \cite{kadir2018designing}, and a lack of knowledge regarding cobots by SMEs \cite{boucher2022smes}. These findings illustrate the difficulty of successfully integrating cobots into existing manufacturing processes.

\subsection{Factors for Cobot Integration}

When beginning to integrate cobots into manufacturing facilities, there are a number of factors that must be considered. Existing work has identified the need to better understand work environments such that cobots can safely operate within them \cite{kildal2018potential, malm2019collaborative}. Maintaining safe operation requires consideration of factors such as crossover between cobot and worker work zones, cobot handling of objects, and cobot movement speeds, as these can create unsafe conditions for operation \cite{bi2021safety}. Furthermore, certain cobot actions (\textit{e.g.}, handling hazardous materials, moving quickly, or moving unintuitively) can create non-collaborative environments. These examples show that collaboration is dictated in part by a given task and is not inherent to the application of cobots themselves \cite{guertler2023robot}.

Once a task is selected and initial workspace factors have been considered, additional interaction considerations must be made. Integrators must consider the ways individuals will interact with cobots to complete a task and utilize their knowledge of a cobot's capabilities to develop a collaborative process that optimizes operator needs and task outcomes\cite{grahn2014benefits, simoes2020factors}. To make a process collaborative, existing work has documented a set of guiding considerations \cite{malik2021reconfiguring}, such as workspace configuration, ergonomic impact, types of interaction and collaboration that occur between the operator and robot, and understandability of cobot actions to the operator.

Task scheduling is well explored in automation \cite{lewandowski2020automated, yin2018tasks, wang2019task}, and cobots provide new variables that integrators need to consider, as they pose new ways of dividing, sharing, and collaborating on tasks between the worker and cobot based on the type of interaction \cite{christiernin2017describe}. While algorithms and approaches for addressing this challenge exist \cite{sadik2017flow, tsarouchi2017human}, it is important to consider the ways in which a cobot can assist the human operator more directly, as it is commonplace for operators to adjust their own workflows to work with cobots \cite{wurhofer2015deploying}. While cobots can improve an operator's physical working conditions \cite{peshkin1999cobots, cardoso2021ergonomics}, this capability is dependent on which tasks are selected for the cobot and operator to perform. It is important to consider the operator's preferences and trust between the operator and cobot, as these are important factors in determining resulting task performance \cite{kopp2021success}.

\subsection{Integration Frameworks and Approaches}

There are many key factors that need to be considered when approaching cobot integration. One of these factors is the selection of candidate workcells and processes for cobot assistance. This step can be completed by identifying any manual processes that may be a bottleneck to other processes \cite{cohen2019strategic}, or through analyzing return on investment over long-term usage \cite{gil2017integration}. Another step that must be completed is the configuration of an effective workcell (\textit{i.e.}, developing a simple, modular, and safe design for workers) \cite{malik2021reconfiguring} while taking into account productivity \cite{gil2017integration} and interactions between the cobot and worker \cite{malik2021reconfiguring}. These workcell designs need to consider the potential for the cobot to work in parallel with the worker, either by having the cobot work in a separate area of the cell or by collaborating with the worker directly \cite{andronas2020design}. Additionally, an appropriate cobot must be selected for integration based on the context in which it will be situated. Prior work has investigated how to make this decision, based on the requirements of the task, the properties of the robot, and its potential performance \cite{cohen2019strategic, ghorabaee2016developing}. 

Several of the above steps have been encapsulated by the National Institute of Standards and Technology (NIST) within their set of guidelines for cobot integration, based on discussions with robotics experts \cite{horst2021best}. In their work, they present several concrete methods for identifying candidate workcells for integration, metrics for selecting a cobot, and metrics for determining the viability of a given integration plan. Overall, NIST provides several steps to begin the integration process, as well as metrics and considerations to use in the decision-making process. Other work has explored a method of integration which begins at a general level, by understanding the task context, and then considers specific elements such as the workcell, cobots and other machines, and, finally, the workers \cite{djuric2016framework}. 

However, recent work has acknowledged the technological focus that exists in prior approaches to cobot integration, as well as the recent shift towards incorporating a socio-technical perspective that considers the worker and cobot a partnership rather than as individuals \cite{adriaensen2022teaming}. While these technological approaches have defined many important factors and methods for integration, they fall short of incorporating both manufacturer and worker considerations while also demonstrating system feasibility.
\section{Approach}

\begin{figure*}
  \centering
  \includegraphics[width=\linewidth]{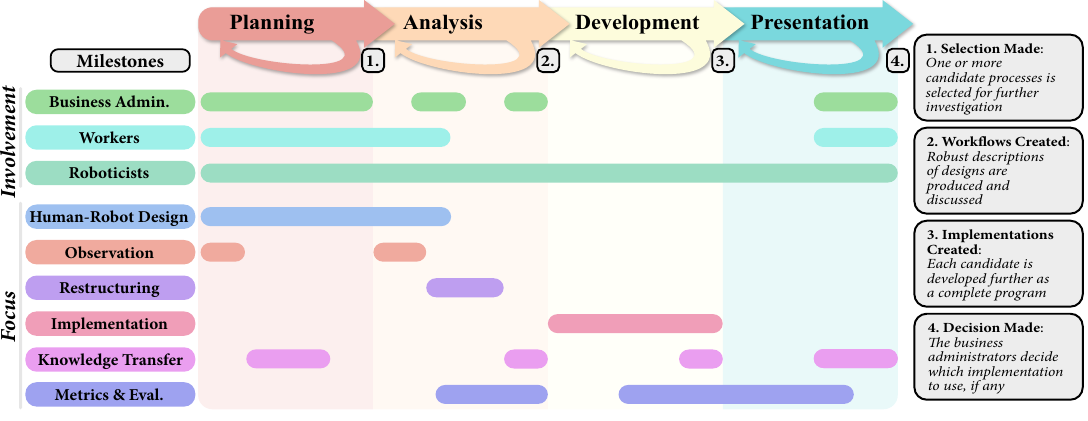}
  \caption{An overview of the four-phase approach including the individuals, foci, and milestones involved at each phase.}
  \label{fig: Collaboration Overview}
\end{figure*}

We propose a four-phase approach to help businesses understand the costs, outcomes, and implications of cobot integration in order to make better-informed decisions (see Figure \ref{fig: Collaboration Overview}). These four phases are \textit{planning} (\textit{i.e.}, understanding the context of cobot intervention), \textit{analysis} (\textit{i.e.}, defining the roles of the human and robot), \textit{development} (\textit{i.e.}, creating a new workflow involving the human and robot), and \textit{presentation} (\textit{i.e.}, gathering and presenting relevant information to the collaborator). These four phases provide a pathway to the integration process that business administrators can utilize to understand the effects of integrating a cobot into their workflows before making major commitments.

\subsection{Planning Phase}

The initial \textit{planning phase} of our approach attempts to develop an understanding of existing workflows and allows the roboticist to ground their expertise within the context of these processes. This approach builds on the ideas of contextual inquiry \cite{beyer1999contextual}, where observations and interviews are combined in order to develop a thorough understanding of a conceptual space. Thus, this phase requires the involvement of all three parties (\textit{i.e.}, business administrators, workers, and roboticists), and places an initial focus on applying the roboticist's understanding of HRI through observations and discussions. This initial understanding may be achieved through on-site visits and tours, or a series of discussions regarding existing processes and workflows currently completed by human workers alone. Within these discussions, it is important to identify contextual factors including the workspace layout, spatial constraints, and resources required for a specific process including labor and parts. Additionally, it is important to identify opportunities for collaborative assistance by the cobot. While there may be several tasks or subtasks that can benefit from cobot involvement, a discussion of worker and business preferences (\textit{e.g.}, reduction of undesirable work or optimization of critical tasks) will guide which tasks are most appropriate. For manufacturers with limited prior knowledge of cobots, this phase helps to ground any potential ideas for integration based on the realistic capabilities of cobots and helps to establish expectations of the impact a cobot could have. Follow-up discussions should occur as often as necessary for the manufacturer to gain a sufficient understanding of cobot integration and to agree upon the best possible task candidates for cobot intervention. By the end of the planning phase, manufacturers should have a cursory understanding of what cobot integration requires and yields, and roboticists should have one or more candidate tasks they can begin to analyze in the next phase.

\subsection{Analysis Phase}

Once the planning phase is complete, the roboticist can begin gathering data on the existing human-only work process for review in the \textit{analysis phase}. This phase will initially require the involvement of all three parties as the phase focuses on data collection. The gathered data may include blueprints of the work environment and its configuration, videos of process execution, timetables of task steps, subtask dependencies, component schematics, or other forms of data that describe the work process in fine detail. After data collection, the involvement of the worker and business administrators is lessened. From here the roboticist uses the collected data to concretely understand the environment and existing workflows so they may be restructured. The restructuring process initially involves analyzing the workflow, which can be done through methods such as hierarchical task analysis \cite{stanton2006hierarchical}, a method used to break a task down into goals and subgoals to understand its operation. Once a given task is understood at a granular level, the roboticist can complete the restructuring process by dividing the overarching goal into subtasks for the worker and the cobot based on worker preferences, robot capabilities, and overall optimization of the task. This process may place particular emphasis on limiting human or cobot idle time to allow for optimal efficacy of the human-cobot team. However, it is important that this process of assigning subtasks leverages the roboticist's knowledge of cobots and their capabilities, and incorporates principles of human-robot interaction and ergonomics. This practice allows the roboticist to ensure that the cobot acts as a effective collaborator, assists the operator in a safe manner, and improves overall task performance \cite{paliga2022human, simoes2022designing, cardoso2021ergonomics}. Given the unique stakeholder preferences that need to be considered in creating a new human-cobot workflow, continued discussion with the collaborator may be necessary to ensure that desired outcomes are achieved. During this phase, it is important that the roboticist develop an understanding of where and how the cobot can be optimally inserted within the existing workflow. Additionally, the roboticist must be aware of potential failure points caused by limitations in the cobot's capabilities. For example, if a particular component does not have convenient grasp points, manipulation of such a component may be a task better suited for the human worker. At the end of the analysis phase, a new human-robot workflow should be produced and communicated to stakeholders for high-level feedback. Based on this feedback, the roboticist may need to iterate on prior planning and analysis steps.

\subsection{Development Phase}

During the \textit{development phase}, the goal of the roboticist is to operationalize the newly created human-cobot workflow and produce outcome metrics that communicate the workflow's performance. This phase will primarily involve the roboticist, as they initially focus on the implementation by creating a simulation of the new workflow process in software systems such as Unity \cite{bartneck2015robot}, RViz \cite{kam2015rviz}, Webots \cite{michel2004cyberbotics}, or CoFrame \cite{schoen2022coframe}. The purpose of the simulation is to act as a general proof of concept that showcases where and how the cobot operates within the environment and demonstrates the feasibility of the new workflow as the collaborative process is executed. Once the implementation is complete, the roboticist will begin to produce metrics derived from the simulation in tandem with updating the implementation as needed. These metrics should account for the process's cycle time and the robot's idle time and include information about potential safety concerns and their mitigation. Based on the workcell setup and cobot that are utilized within the simulation, a roboticist can begin to approximate the price of components needed to recreate the simulation within the manufacturer's facility. As this integration plan is developed, potential flaws may become evident, thereby necessitating additional stakeholder discussion and iteration on prior completed steps. By the end of the development phase, the roboticist should have a concrete integration plan including the simulated workflow, process outcome metrics, and approximate component costs.

\subsection{Presentation Phase}

In the \textit{presentation phase}, the roboticist synthesizes all information from the prior three phases and discusses it with the business administrators and workers. These data may include the procedure of the new human-cobot workflow, subtask timetables, process performance metrics, equipment or labor costs of integration, and overall profit per produced item. Once gathered, these data can be compared to the existing human-only work process and analyzed to determine the relative costs and benefits of the human-cobot task procedure and hardware installation or any variants that may have been developed. This information can be complex and cumbersome, so the roboticist may develop recommendations based on specific overarching needs and preferences conveyed by collaborators. All information should be formatted for submission to the manufacturing collaborator (\textit{e.g.}, within a presentation or written report) and then discussed to ensure they fully understand the results and have any questions or concerns addressed. Once the presentation phase is complete, the manufacturer should have a thorough understanding of the potential integration plan, its outcomes, and any other information required to make an informed decision about cobot integration within their facility. From this point, additional discussion and iteration on proposed ideas can occur depending on the needs of the business and nature of the collaboration.
\section{Case Study Application}

\subsection{Collaboration Context}

Our team worked with a local SME manufacturer that expressed interest in cobot integration. In working with our collaborator, we took several actions to ensure the privacy and confidentiality of the business as well as individuals with whom we interacted. First, we have omitted the identity of our collaborator in this paper and supplemental documentation. Second, our institution signed a non-disclosure agreement (NDA) with our collaborator, and the research team sought permission from our collaborator to publish the material presented in this paper. Third, we collected information (\textit{i.e.}, video recordings, process information, and feedback) only after verbal consent was obtained (in process analyses) or consent forms were signed (in feedback sessions).

While our collaboration lasted approximately eight months, this time period included iterative modifications to the software tools we used for analysis and simulation. We expect our proposed approach to take less time with a strict minimum of four meetings (\textit{i.e.}, initial discussion, collection of data, overview of implementation, and presentation) and additional meetings as needed depending on the nature of the collaboration. Therefore, we expect our approach to roughly require a time commitment between a few days and a few weeks. More streamlined software tools and organizational commitment can shorten this timeframe to a few hours.

Using our proposed approach, we present a case study of its application with our collaborator. In this section, we present the application of each of the four phases, actions that were taken in each, and feedback provided following our initial collaboration.

\subsection{Planning Phase: Stakeholder Discussions}

Our initial meeting with our collaborator involved touring their facility, discussing business needs and worker preferences, and seeking preliminary opportunities for cobot integration. From this initial meeting, we learned that the business administrators wanted to increase efficiency in their process to fulfill more product orders and saw cobots as a means of meeting this need. Additionally, the administrators expressed interest in having a cobot take over undesirable and messy tasks from the workers, a sentiment that was echoed by the workers themselves. The result of this initial visit helped set expectations with the manufacturer and allowed us to identify several potential areas for a cobot to assist in their process. Following this visit, our research team convened to discuss potential options and scheduled a follow-up meeting with the manufacturer to further discuss the potential of each option. 

During the follow-up visit, we developed a deeper understanding of the various tasks that could potentially benefit from cobot intervention. By observing worker processes, we were able to identify tasks that were repetitive or undesirable to workers. At the end of the visit, we discussed with the administrators which tasks would be best suited for cobot integration given our knowledge of cobots, our understanding of each process's potential for collaboration, the administrator's desired business outcomes, and the preferences of workers. From this discussion, an assembly task was chosen which required the collection of parts, silicone application, rivet fastening, and other subtasks. This particular task involved the assembly of high-volume units, which was a high priority for reduced cycle time and could lead to an increase in their throughput and overall ability to fulfill orders. The application of silicone in the assembly task was seen as extremely messy and therefore undesirable to workers, while also being ideal for the cobot given its ability to perform precise and repetitive motions. As a second candidate, we identified that the cobot could assist further by supplying components and then applying silicone to them. Overall, the selected task was a clear fit for cobot integration and was the focus of further analysis.

\subsection{Analysis Phase: Existing Workflow}

Once a candidate task was identified, we visited the manufacturing facility again to gather detailed information about the specific task. We filmed workers performing the task, to be analyzed later, and asked them questions to clarify the general assembly procedure and variants that are utilized by different workers. Video was collected only after receiving verbal consent from workers and was stored and shared with our collaborator through university-approved digital storage. No employee identifying information was collected through these videos (\textit{i.e.}, verbal information or view of worker faces). In consultation with our Institutional Review Board (IRB), these interactions were not considered research with human participants, as they focused on the manufacturing processes rather than the individuals, although feedback sessions were, as described in \S\ref{sec:feedback}. Additionally, administrators provided schematics for parts and workcell layouts utilized within the assembly process. 

\begin{figure*}[!t]
  \centering
  \includegraphics[width=\linewidth]{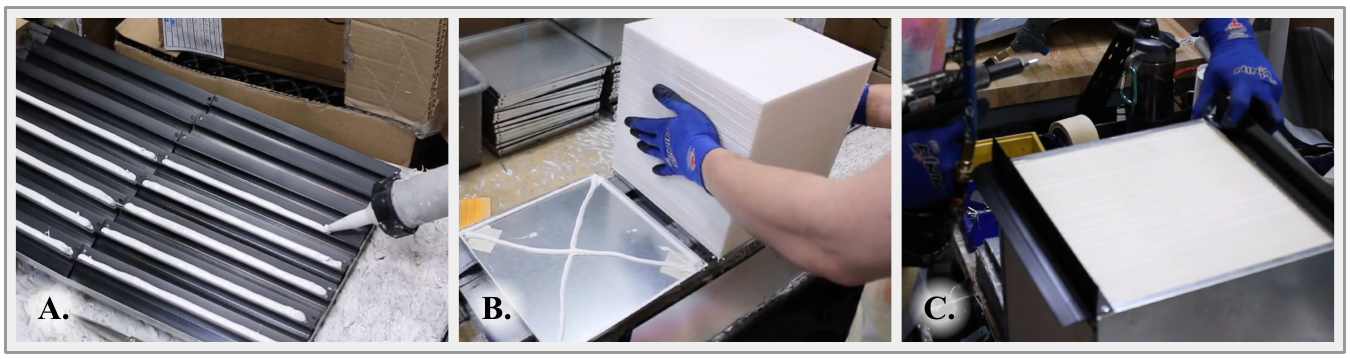}
  \vspace{-4mm}
  \caption{A subset of the steps involved in the manufacturer's assembly process. A. applying silicone to gaskets. B. applying silicone to pans and beginning the assembly process. C. continuing the assembly process by attaching gaskets.}
  \label{fig:human-worker}
  \vspace*{-2mm}
\end{figure*}

We used the collected data to reconstruct the assembly process to understand the steps involved in the procedure, and how it varied between workers. This process is shown in Figure \ref{fig:human-worker}. We first analyzed the video to create a timeline of the assembly process. Some segments included discussions with workers and additional pauses during which workers would provide the camera with specific views of the process or parts. These sections were removed from our timeline reconstruction for accuracy. Additionally, when appropriate, durations of specific steps were averaged between workers. Our reconstructed timeline can be seen in Table \ref{tab:human-process}.

\begin{table}[]
  \centering
  \caption{A timeline of the SME's existing procedure to complete one cycle of their assembly process.}
  \includegraphics[width=\linewidth]{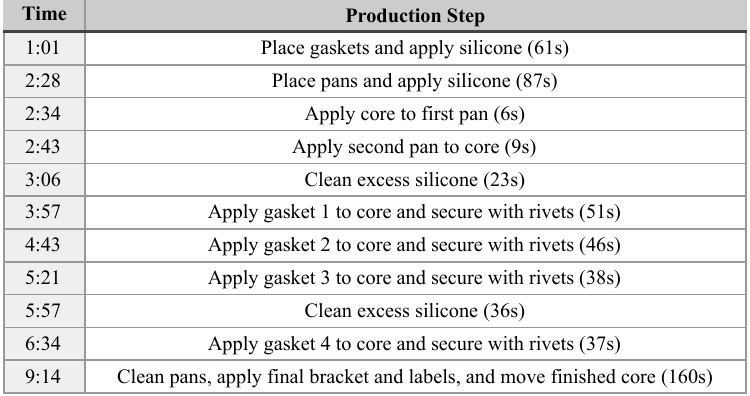}
  \vspace*{-5mm}
  \label{tab:human-process}
\end{table}

Once we reconstructed a timeline of the human-only process, we identified potential subtasks where the cobot could assist the operator in the process. These subtasks were: (1) the cobot applying silicone to components placed by the worker and (2) the cobot picking and placing components, and applying silicone to them. While we had noted a possible delineation of work at the end of the planning phase, this step helped us to formally identify and justify the distribution of subtasks. While both assignments appeared feasible, it was unclear what the optimal selection would be given the differences in benefits to the worker and process, as well as the required component costs. Exploring these different allocations of work allowed us to provide our collaborator with multiple options to consider depending on their budget and business needs. Both options were presented and confirmed to be practical. Using these potential processes, we next created two new timelines to represent the potential human-cobot process accounting for the possibility of one or two workers being assisted by a single cobot simultaneously. These approximate time-tables are used as the basis for the development phase, as they provide a rough outline of what needs to be achieved and in what time frame.

\subsection{Development Phase: Simulation}

\begin{figure}[!b]
  \centering
  \vspace{-2mm}
  \includegraphics[width=\linewidth]{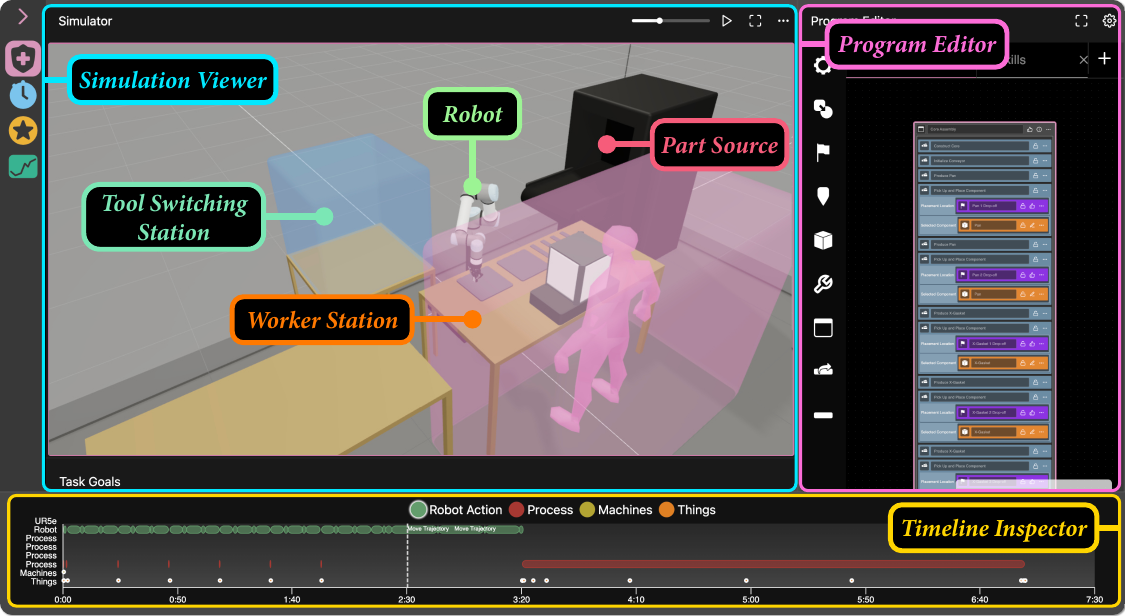}
  \caption{The program and simulated environment created for process 2. The environment captures the robot, a workspace for the operator, a workstation, a location to switch end effectors, and a source for component parts.}
  \label{fig:Simulation 2}
\end{figure}

Based on the approximate timetables that were created in the analysis phase, we created simulations of the new human-cobot workflows using the CoFrame \cite{schoen2022coframe} system. Within the simulation, we initially modeled the workcell of the manufacturer based on the information they provided and adjusted it to reflect the changes brought by integrating a cobot into the space. The created models included walls to visualize spatial constraints, tables where the component preparation and assembly process occurs, a table on which a tool switcher would be placed, and a conveyor to symbolize a location for component pickup (see Figure \ref{fig:Simulation 2}). After modeling the environment, we added models for the cobot, a UR5e robot arm, models for the components that are required within the assembly process of a single product unit (\textit{i.e.}, a core, pans, and gaskets), and defined regions of the workspace that the operator would work in. With the work environment, cobot, and components modeled, we next began to create simulations of the cobot and worker tasks.

\subsubsection{Process 1: Silicone Application Only}

The first process we created in simulation involved the worker placing components, the cobot applying silicone to those components, and the worker completing the remaining steps in the assembly process. A simplified version of the final timeline for this workflow can be seen in Table \ref{tab:process-1-timeline}, and reflects the cobot's ability to assist two workers in parallel. Within our simulation, we defined locations for components to be placed and waypoints for the cobot's end effector to follow as it applied silicone. The cobot's end-effector speed was optimized for efficiency and safety based on feedback from CoFrame's review panel. The simulation also included processes to simulate the worker assembling components to recreate an entire production cycle and produce performance metrics that accurately captured the entire human-cobot workflow. These performance metrics included cycle time, cobot idle time, and wear-and-tear cost.

\begin{table}[b]
  \centering
  \vspace*{-3mm}
  \caption{A timeline of process 1 including the steps and timing for a cobot to assist one or two workers in applying silicone.}
  \label{tab:process-1-timeline}
  \includegraphics[width=\linewidth]{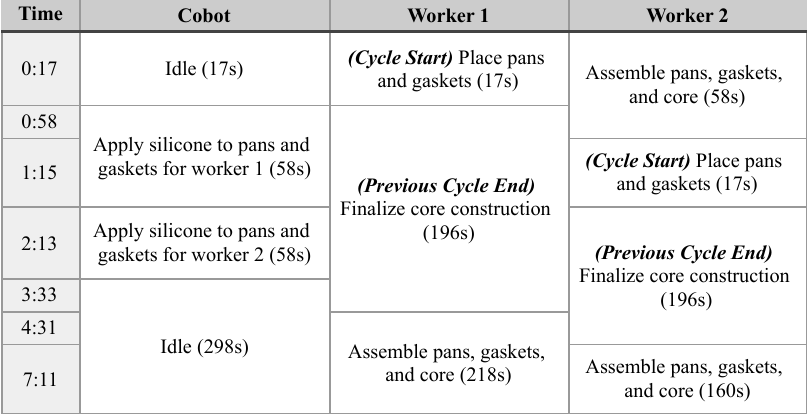}
\end{table}

\subsubsection{Process 2: Silicone and Pick-and-Place Workflow}

Our second simulated workflow involved the cobot picking up, placing, and applying silicone to components, and the worker completing the remainder of the assembly process. Similar to our first simulation, locations and waypoints were defined for components and the cobot's end-effector path. However, in this simulation, after placing components, the cobot navigates to a designated tool switcher zone to exchange its gripper for a silicone dispenser. The cobot's end-effector speed was calibrated based on feedback from CoFrame and the performance metrics that it generated. While this workflow offloads the additional task of placing components from the worker to the cobot, and resultingly reduces cobot idle time, this process still allows the cobot to assist two workers in parallel. Table \ref{tab:process-2-timeline} shows the timeline for this workflow, and Figure \ref{fig:Simulation 2} illustrates the simulation setup.

\begin{table}[!t]
  \centering
  \caption{A timeline of process 2 including the steps and timing for a cobot to assist one or two workers in gathering components and applying silicone.}
  \includegraphics[width=\linewidth]{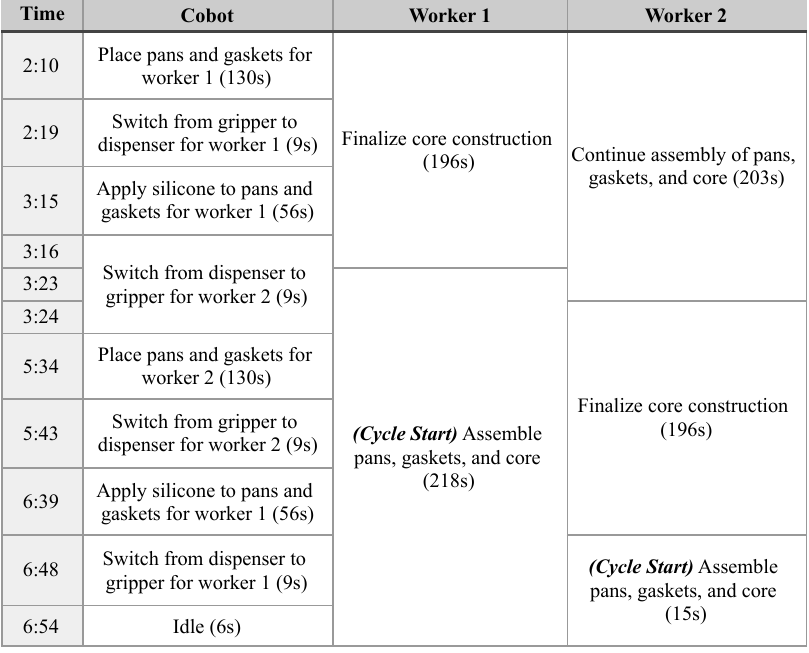}
  \label{tab:process-2-timeline}
  \vspace*{-7mm}
\end{table}

\subsubsection{Equipment Costs}

As the final step of this phase, we obtained cost estimates for the cobot and other materials required for integration based on our simulations. These estimates helped to develop requirements to reconstruct the simulations within the manufacturer's facility. Cost estimates were obtained through discussions with a robotics vendor. While not included within our simulations, we also obtained cost information for a range extender that would allow the cobot to move between two workstations. If the range extender were to be utilized, the manufacturer could use a single cobot to assist two workers in parallel with minimal changes to their existing setup. Alternatively, the same benefits may be achieved by reconfiguring the workspace such that a single cobot could directly access two workspaces. These hardware options were included to allow for greater flexibility in achieving preferred outcomes and allowed us to present multiple options to the manufacturer so they could select one based on their needs and constraints.

\subsection{Presentation Phase: Reporting Results}

Using the performance metrics that were generated from our simulations and the cost information that we obtained for each workflow, we created a complete write-up to describe both processes, their associated costs, and the trade-offs of each. We presented these findings to the manufacturer, describing in detail each plan, showing them the simulations and output metrics of each, and answering any remaining questions they had.

\subsubsection{Synthesizing Results}

\begin{figure}[t]
  \centering
  \includegraphics[width=\linewidth]{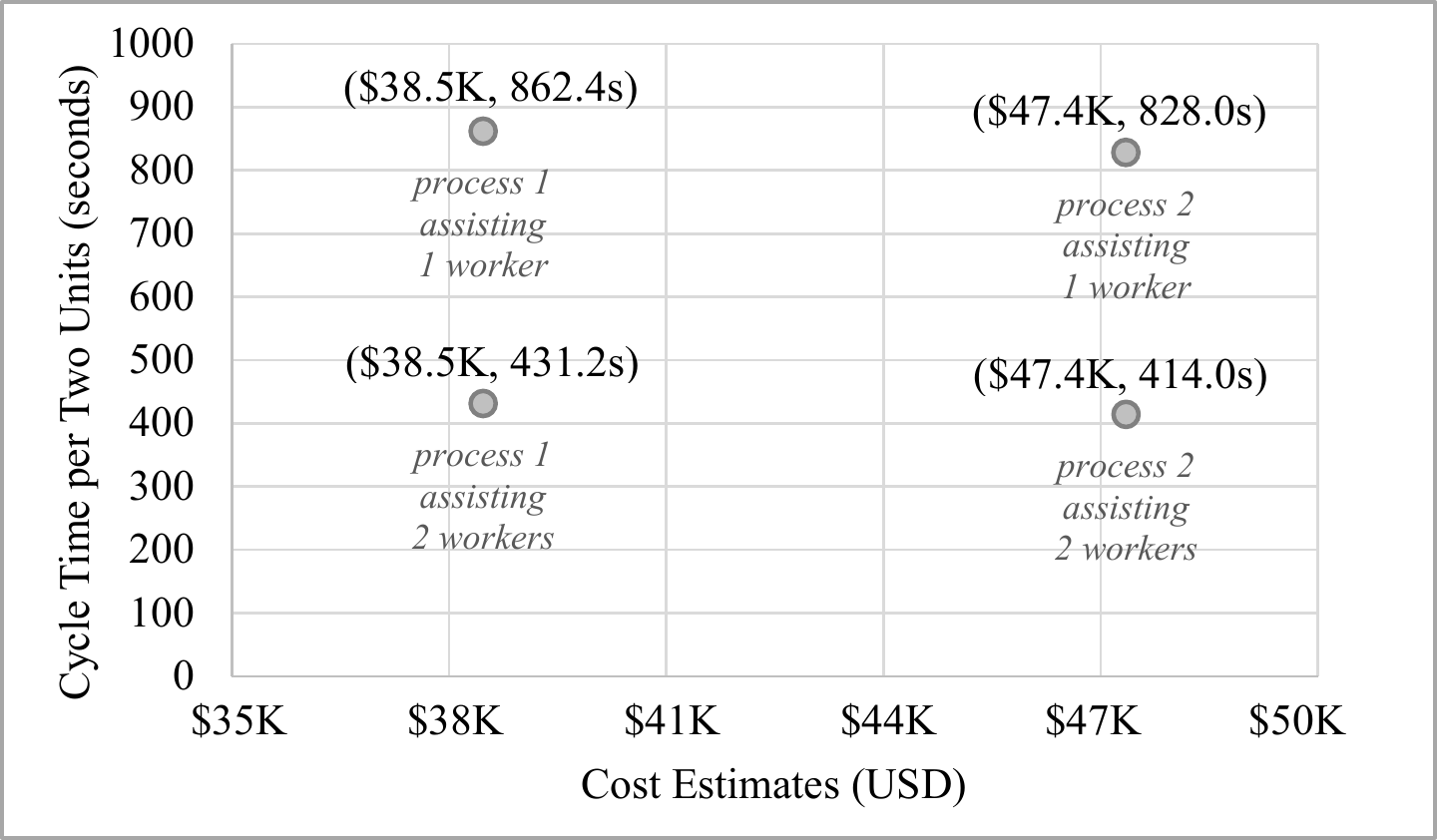}
  \caption {This plot showcases the costs of our processes and their cycle times to assemble two units when one or two workers are assisted by the cobot.}
  \label{fig:CostVSTime}
  \vspace*{-5mm}
\end{figure}

Based on our analysis of the original human-only workflow, the total cycle time was 9:14 minutes. By comparison, our simulation of processes 1 and 2 yielded cycle times of 7:11 minutes and 6:54 minutes, respectively. As a result, processes 1 and 2 would reduce total cycle time by approximately 22\% and 25\% respectively. However, the manufacturer would need to weigh the improvement these reduced cycle times provide against the total costs of integration. For process 1, our estimated integration cost was \$38,470.00 USD, but resulted in more idle time by the cobot. Process 2's increased task assignment reduced this idle time, but required additional hardware and would cost an estimated \$47,350.00 USD to be integrated into the facility. Additionally, both workflows would incur some level of wear-and-tear cost from regular operation. According to our simulation, this cost would be negligible for process 1 but would be \$0.02 USD per cycle for process 2. Both workflows produced clear benefits from a business perspective, but process 2 was able to fulfill worker preferences by offloading an undesirable task (\textit{i.e.}, applying silicone to components) to the cobot, although at a much higher cost (shown in Table \ref{tab:Metrics} and Figure \ref{fig:CostVSTime}).

\begin{table}[t]
  \centering
  \caption{This table shows several metrics comparing both proposed processes and their direct differences.}
  \includegraphics[width=\linewidth]{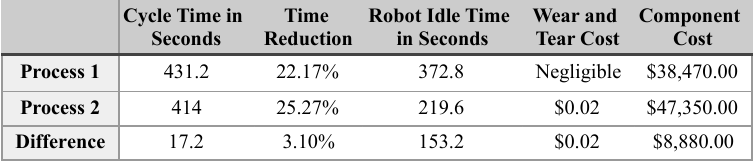}
  \vspace*{-7mm}
  \label{tab:Metrics}
\end{table}

\subsubsection{Presenting Results}

During our final meeting with our collaborator, we presented all the results uncovered throughout our collaboration. We reviewed the overall steps we had taken over the span of our several-month partnership, focusing on the factors that motivated the need for cobot intervention within our collaborator's facility and highlighting the decisions made along the way to progress toward the final proposals. We discussed the process required to create our simulations, including the simulation environment (\textit{i.e.}, CoFrame), our model of the manufacturer's workspace, and the cobot we utilized within the simulations. This overview helped to familiarize our collaborator with the simulations before providing a full demonstration. Next, we provided a detailed demonstration of each workflow simulation along with simplified timelines to illustrate each process. When showing process 2, we provided an overlay video of the human work process to clearly visualize the worker's and cobot's coordinated effort. These demonstrations provided an intuitive and compelling view of the cobot's effectiveness. For each of the simulations, we discussed the required integration costs and the performance metrics they each achieved. While the ultimate decision of whether and how to integrate a cobot belonged to our collaborator, we attempted to provide a clear understanding of the benefits of each workflow and how they compared to one another. Following our presentation, our collaborator had several questions regarding potential next steps and practical considerations if they were to pursue either option further. We addressed these questions and offered further support, should the need arise. 

\subsection{Feedback Session}\label{sec:feedback}

Following our final presentation with our SME collaborator, we held one additional discussion session to receive feedback on our presented work and four-phase approach. Within this session, we sought feedback from two manufacturing engineers and a manufacturing engineer manager. While we expressed interest in involving production workers in this session, we were unable to due to their work schedules. At the beginning of the session, consent forms were read and signed by all participants. During the session, we briefly presented our four-phase approach and conducted a semi-structured interview with the participants as a group. The session was conducted over Zoom and recorded. We transcribed the interview and conducted a thematic analysis of the interview data, which revealed three primary findings discussed below.

First, we learned that our approach was able to capture worker needs at all levels. The manager emphasized that production workers would be very pleased if an undesirable task (\textit{i.e.}, messy silicone caulk application) were offloaded to a cobot partner. He further emphasized that the use of cobots in this process would not eliminate the worker, but rather make their work more efficient. Additionally, the engineers reported that their needs were met by providing relevant information including robot motion sequences, spatial constraints, and comparisons to the currently manual process. Finally, the manager reported that our collaboration answered many of their questions about cobot application.

Second, our work acted as an effective proof of concept for our collaborator. The manager expected that work toward cobot integration would occur in phases and that the first phase would be determining feasibility. The manager stated, \textit{``I think this project really helped us answer the question, 'can we do this?', right? 'Is this a good application?' I think that really helped us.''} Specifically, our collaborators expressed that the simulations illustrated the ``big picture'' of what such a workflow could resemble and allowed for a clear comparison to the existing processes. These sentiments convey how our proposed process can answer initial critical questions that may be stumbling block for many businesses in evaluating whether cobot integration is appropriate in their facilities.

Finally, we learned that, although our case study provided our collaborator with an understanding of feasibility toward making informed decisions, implementing our presented solutions would require additional action. Our collaborators explained that management approval for such a project would require information including comparisons across available cobot systems, observation of these systems in actual production, and generation of supplier lists. Then, resulting elements such as exact workcell layouts, equipment specifications, and cost estimates can be proposed for budget approval. Although some of these implementation steps extend beyond what our approach encapsulates, they can clearly benefit from and build upon the results generated utilizing this approach.

\section{Discussion}

Through our collaboration with an SME manufacturer, we demonstrated the steps required to apply our proposed approach, develop new human-cobot workflows, and assist a business in making an informed decision regarding cobot integration. This experience offered critical insights into the information that a business needs to make informed decisions about integration and that roboticists require to assist that business. These insights show that our approach can begin to answer our initial questions regarding complementary cobot intervention, worker preferences, and successful cobot integration. First, our approach can offer a deep understanding of the manufacturer's workflow and creates solutions that utilize cobots' collaborative capabilities. Further, our approach can enable roboticists to apply principles of collaborative robotics, ergonomics, and HRI to develop effective workflows that meet worker preferences. Finally, our approach can help satisfy the needs of the larger business without sacrificing the prior two goals. This intersection of proper cobot utilization, consideration of worker preferences, and optimization of business goals is critical to cobot integration.

While our case study illustrates the utility of our proposed approach, certain elements warrant further exploration. First, our proposed human-cobot processes were developed only in simulation. Given that simulations cannot accurately reflect real-world performance, further \textit{in situ} evaluations will likely produce additional insights into the effectiveness of our approach. Additionally, while our case study proceeded in a relatively linear fashion, further evaluation of our approach's iterative capacity (\textit{i.e.}, the ability to iterate between phases rather than only within individual phases) may help us to understand its ability to handle complex manufacturing processes. Furthermore, to make the integration process more accessible for SMEs, future work can seek to develop tools and frameworks that support SMEs in their endeavor to make informed decisions about cobot integration.
\section{Conclusion} 
While cobots hold great potential for use within SME manufacturing facilities, SMEs may not have the required knowledge to fully utilize the collaborative capabilities of cobots or understand the implications of their integration. Effective integration requires consideration of human worker needs and preferences, proper utilization of cobot strengths, and improved performance of manufacturing processes. In this paper, we presented a four-phase approach to the cobot integration process and illustrated its use through our experience with an SME manufacturer. Integrating cobots into existing manufacturing workflows requires extensive knowledge of cobots, their capabilities, HRI principles, and a deep understanding of the manufacturer's needs. Our approach facilitates this deep understanding, assists in the development of effective integration proposals, and supports SME manufacturers in making informed decisions about cobot integration within their facilities.

\begin{acks}
We would like to thank David Kwon for his assistance in gathering cost information, the reviewers for their insightful feedback, and our manufacturing collaborator for their contribution to our case study. This work received financial support from a Piloting Research Innovation \& Market Exploration (PRIME) grant from UW–Madison’s Discovery to Product (D2P) and National Science Foundation Awards 1822872, 1925043, 2026478, and 2152163.
\end{acks}

\balance
\bibliographystyle{ACM-Reference-Format}
\bibliography{references}

\end{document}